\def\year{2020}\relax

\documentclass[letterpaper]{article} 
\usepackage{aaai20}  
\usepackage{times}  
\usepackage{helvet} 
\usepackage{courier}  
\usepackage[hyphens]{url}  
\usepackage{graphicx} 
\usepackage{graphicx}  
\usepackage{amsmath}
\usepackage{amssymb}
\usepackage{xspace}
\usepackage{url}
\usepackage{xcolor}
\newcommand{\causalstate}{\ensuremath{\xi}\xspace}

\newcommand{\CausalStateSet}{\ensuremath{\mathbf{\Xi}}\xspace}
\newcommand{\plc}{\ell^-}
\newcommand{\PLC}{\mathtt{L}^-}
\newcommand{\flc}{\ell^+}
\newcommand{\FLC}{\mathtt{L}^+}
\newcommand{\fhorizon}{h^+}

\newcommand{\site} {\mathbf{r}\xspace}
\newcommand{\point} {(\site, t)\xspace}
\newcommand{\radius} {R}
\newcommand{\neighborhood} {\eta}
\title{Spacetime Autoencoders Using Local Causal States}

\author{Adam Rupe\thanks{atrupe@ucdavis.edu} and
James P. Crutchfield\thanks{chaos@ucdavis.edu}\\
Complexity Sciences Center and Department of Physics and Astronomy,
University of California Davis\\
One Shields Avenue, Davis CA. 95616
}

\begin{document}

\maketitle

\begin{abstract}
Local causal states are latent representations that capture organized pattern
and structure in complex spatiotemporal systems. We expand their functionality,
framing  them as spacetime autoencoders. Previously, they were only considered
as maps from observable spacetime fields to latent local causal state fields.
Here, we show that there is a stochastic decoding that maps back from the
latent fields to observable fields. Furthermore, their Markovian properties
define a stochastic dynamic in the latent space. Combined with stochastic
decoding, this gives a new method for forecasting spacetime fields.
\end{abstract}

\section{Introduction}

Physics-based representation learning is a key emerging tool for analyzing
high-dimensional nonlinear systems~\cite{Will20a,Brun20a}, and one with a long history
in dynamical systems \cite{Crut87a}. Nonlinear systems often require additional
tools beyond computer simulation to extract actionable insight from the complex
behaviors they exhibit. There are two main aspects to
physics-based representation learning. The first is
\emph{dimensionality-reduction}: Learn a low-dimensional representation of a
behavior---such as coherent structures in fluid
flows~\cite{Holm12a,Peac13a}---that provide a more human-interpretable
accounting for the full system's behavior. The second is \emph{generative
modeling}: Provide a modeling alternative to numerical simulation of the
equations of motion that can be applied directly to data. This is particularly
helpful if the proper equations of motion are unknown. For high-dimensional
systems the data-driven models are typically computationally less expensive than direct
numerical simulation.


For learning representations of high-dimensional nonlinear dynamics, several
approaches have been introduced. \emph{Proper Orthogonal Decomposition}
(POD)---rather similar to \emph{Principle Component Analysis} (PCA)---is a
canonical method for fluid flows \cite{Holm12a,Rowl17a}. POD modes provide a
low-dimensional latent representation that can give interpretable insights into
a flow's large-scale organization. Flow equations of motion may also
incorporate a set of POD modes, through Galerkin projection, giving a truncated
set of ODEs as a generative model that is less expensive to simulate than the
full Navier-Stokes PDE. Like PCA, the POD modes are linear latent
representations. For that matter, PCA is equivalent to a linear
autoencoder~\cite{Bald89a}. Representation linearity substantially restricts
the complexity of flow structures that can be modeled appropriately.

Addressing this, the Perron-Frobenius and Koopman operators recently gained
popularity as nonlinear generalizations for spectral (modal) analysis of
high-dimensional dynamical systems~\cite{Froy09a,Mezi13a,Klus19a}. As with
nonlinear autoencoders, these operators' modes are typically used for nonlinear
dimensionality-reduction. 

Similarly, the Koopman operator~\cite{Alex20a} and autoencoders~\cite{Hern18a}
are now used for generative modeling which, in the dynamical systems setting,
is a form of \emph{predictive forecasting}. As an aside, reservoir computing
was shown to be effective for predictive modeling~\cite{Path18a}. It is
somewhat analogous to Koopman operator approaches, as the system dynamics are
learned in a higher-dimensional latent space.

Success with operator-approximation methods turns on a
fortuitous matching of their chosen (or inherent) function-basis dictionary and
a system's emergent structures. The reality, though, is that spatially-extended
nonlinear systems generate a diverse set of complicated
organizations---vortices, target patterns, dislocations, and the like. In
short, these emergent structures are not easily or naturally modeled in terms
of known or numerically-approximated spatially-global function bases.
Recent work employs nonlinear autoencoders~\cite{Lusc18a,Mard18a,Otto19a} 
as a means to bypass explicit dictionary choices. Issues still persist with these methods, however; 
finding the ``best'' finite-dimensional operator approximations for a given application 
remains an open problem.  

Local causal states are yet another tool for physics-based representation
learning. They start from a markedly different conceptualization of latent
space, however, with the promise of learning the more complex emergent patterns
generated by nonlinear pattern-forming systems. Recall that reducing dimension
through spectral decomposition implicitly assumes algebraic representations that
are spatially global and spatially coherent. In contrast, local causal states
are spatially-local latent representations that are learned at each point in
spacetime. As such, they are better adapted to capture structures that
self-organize from local interactions governing the dynamics---structures that
consist of many localized or ``coherent'' substructures \cite{Rupe18a,Rupe19a}.
Crucially, since the representations are learned locally, the latent space
shares the same coordinate geometry as the observable spacetime fields. This
adds, among other benefits, a helpful visual interpretability. The following
formally connects local causal states and spacetime autoencoders and presents
preliminary results for predictive forecasting using them.

\section{Local Causal States}
%

%
The local causal states are part of the \emph{computational mechanics} framework~\cite{Crut12a}, which learns nonparametric models of dynamical systems in an unsupervised fashion using the \emph{causal equivalence relation}:
\begin{align*}
\mathrm{p}_t \sim_\epsilon \mathrm{p}_{t^\prime}
  \iff \Pr(\mathrm{Future_t} | \mathrm{p}_t)
  = \Pr(\mathrm{Future_{t^\prime}} | \mathrm{p}_{t^\prime})
  ~.
\end{align*}
The induced equivalence classes over pasts $\{\mathrm{p}_t\}$  are a system's
\emph{causal states}---the unique minimal sufficient statistic of the past for
optimally predicting the future.

\begin{figure}[ht!]
\centering
\includegraphics[width=0.5 \textwidth]{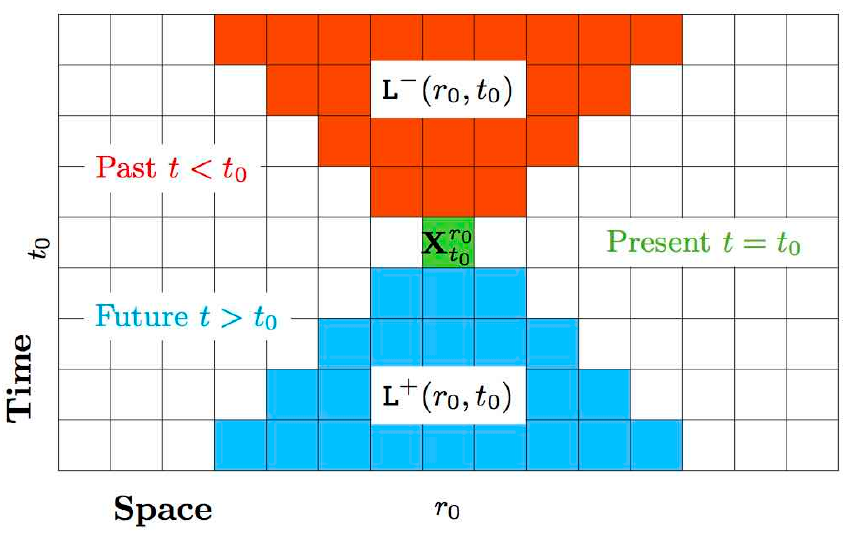}
\caption{Lightcones: past lightcone $\PLC(r_0, t_0)$ (red) and future lightcone
	$\FLC(r_0, t_0)$ (blue) shown for spacetime point $X(r_0, t_0)$ (green) for
	a system with radius $\radius=1$ interactions, such as the map
	lattice example shown below.
	}
\label{lightcones}
\end{figure}

For spatiotemporal systems, \emph{lightcones} are local features that represent pasts
and futures; see Fig.~\ref{lightcones}. Lightcones capture the history and propagation of local interactions in the system through space and time. A lightcone \emph{configuration} is an assignment of observable values to the lightcone templates shown in Fig.~\ref{lightcones}. Two past lightcone configurations $\ell^-_i$ and $\ell^-_j$ are causally equivalent if they have the same
conditional distribution over future lightcones $\mathrm{L}^+$:
\begin{align*}
\ell^-_i \sim_\epsilon \ell^-_j
  \iff \Pr(\mathrm{L}^+ | \ell^-_i) = \Pr(\mathrm{L}^+ | \ell^-_j)
  ~.
\end{align*}
The resulting equivalence classes are the system's \emph{local causal
states} \cite{Shal03a}. They are the unique minimal sufficient statistic of
past lightcones for optimally predicting future lightcones.

\subsection{Encoding}

The lightcone equivalence relation can be recast as a function that generates
the causal equivalence classes---the \emph{$\epsilon$-function} $\epsilon :
\ell^- \mapsto \xi$ maps from past lightcone configurations $\ell^-$ to local
causal states $\xi$. This local mapping from observables to their corresponding
latent local causal state representation is central to using and interpreting
our method. Specifically, $\epsilon\bigr(\ell^-)$ is applied in parallel to all
points $(\vec{r}, t)$ in a spacetime field $X$, mapping the entire field to its
associated \emph{local causal state field} $S = \epsilon(X)$. Every feature
$X(\vec{r},t)$ is mapped to its latent variable (local causal state) via its
past lightcone $\xi = S(\vec{r}, t) = \epsilon\bigr(\ell^-(\vec{r}, t)\bigl)$.
One  result is that the global latent spacetime field $S$ maintains $X$'s
coordinate geometry such that $S(\vec{r}, t)$ is the local latent variable
corresponding to the local observable $X(\vec{r}, t)$. (This is markedly unlike
neural network autoencoders.) The shared geometry of the observable space $X$
and the latent space $S$ facilitates extracting physical features of $X$ from
special (e.g., algebraic) properties in the corresponding spacetime region in
$S$.

For real-valued systems, such as the map lattice analyzed shortly, local causal
state inference requires a discretization to empirically estimate
$\Pr(\mathrm{L}^+ | \ell^-)$~\cite{Goer12a}. Rather than discretize observable
space, we discretize in the lightcone feature space using K-Means to cluster
lightcones using distance metric:
\begin{align}
\mathrm{D}_{\mathrm{lc}}(\mathbf{a}, \mathbf{b})
  \!\equiv\! \sqrt{(a_1 \!-\! b_1)^2 + \ldots + \mathrm{e}^{-\tau d(n)}(a_n
  \!-\!  b_n)^2}
  ,
\label{lcdist}
\end{align}
where $\mathbf{a}$ and $\mathbf{b}$ are flattened lightcone vectors, $d(n)$ is the temporal depth of the lightcone vector at index $n$, and $\tau$ is the temporal decay rate ($1/\tau$ can be thought of as a coherence time).

Two past lightcones are considered $\gamma^-$-equivalent if they are placed
into the same cluster $C^-$ by the distance-based clustering:
\begin{align}\gamma^-(\plc_i)
\!=\! \gamma^-(\plc_j) \!\iff\!
  \plc_i \!\in C^- \; \mathrm{and} \; \plc_j \!\in C^-
  ~.
\end{align}
Similarly for future lightcones. 

This allows us to build
empirical distributions over lightcone-clusters from simple counting. Two past
lightcone clusters are considered $\psi$-equivalent if they have
(approximately) the same empirical predictive distributions: 
\begin{align}
\psi(C^-_i) = \psi(C^-_j) \! \iff \! \Pr(C^+ | C^-_i) \! \approx\! \Pr(C^+
| C^-_j)
~.
\end{align} 
This approximates the $\epsilon$-function as: 
\begin{align}
\epsilon(\plc) \approx
\psi\bigl( \gamma^-(\plc) \bigr)
~.
\end{align}

\subsection{Decoding}

Previously, when performing coherent-structure segmentation the
$\epsilon$-function \emph{encoded} observable spacetime fields $X$ to a
corresponding latent local causal state field $S$~\cite{Rupe18a,Rupe19a}. Such
dimensionality-reduction identifies coherent structures pointwise in observable
spacetime fields. We now introduce, for the first time, the
$\epsilon^{-1}$-function---a stochastic \emph{decoding} that maps from latent
spacetime fields $S$ to \emph{reconstructed} observable fields $\overline{X}$.

Each local causal state $\causalstate$ in $S$ is defined by its predictive distribution $\Pr(\FLC | \causalstate)$, since every past lightcone configuration $\plc_i$ in $\causalstate$ has, by definition, the same predictive distribution: $\Pr(\FLC | \plc_i) = \Pr(\FLC | \causalstate)$, for all $\plc_i \in \causalstate$.
The decoding $\overline{X} = \epsilon^{-1}(S)$ is performed by sampling the
distributions $\Pr(\FLC | \causalstate)$ for each $\causalstate$ in $S$ as
follows. For each spacetime coordinate $\point$:
\begin{enumerate}
\setlength{\topsep}{-2pt}
\setlength{\itemsep}{-2pt}
\setlength{\itemindent}{4pt}
\item Retrieve the local causal state $\causalstate = S\point$;
\item Sample a future cluster $C^+$ from $\Pr(C^+ | \causalstate)$;
\item Retrieve the centroid $\overline{\flc}$ of $C^+$; and then
\item Place $\overline{\flc}$ in $\overline{X}$, with base at $\point$.
\end{enumerate}

Since we use K-Means to cluster both past and future lightcones, we take the
centroid $\overline{\flc}$ as a representative real-valued future lightcone for
cluster $C^+$. Due to the spacetime extent of nontrivial future lightcones,
each point $\overline{X}\point \in \overline{X}$ makes predictions from
several local causal states. In fact, for future-lightcone depth $\fhorizon$
a prediction will be made from each point in its $\fhorizon$-depth past
lightcone. The ultimate prediction for $\overline{X}\point$ averages these
predictions, with the same time-exponential weighting used in the lightcone
metric of Eq.~(\ref{lcdist}). That is, predictions made further out in time are
discounted exponentially compared to more recent predictions.

\begin{figure*}[ht!]
\centering
\includegraphics[width=1.0 \textwidth]{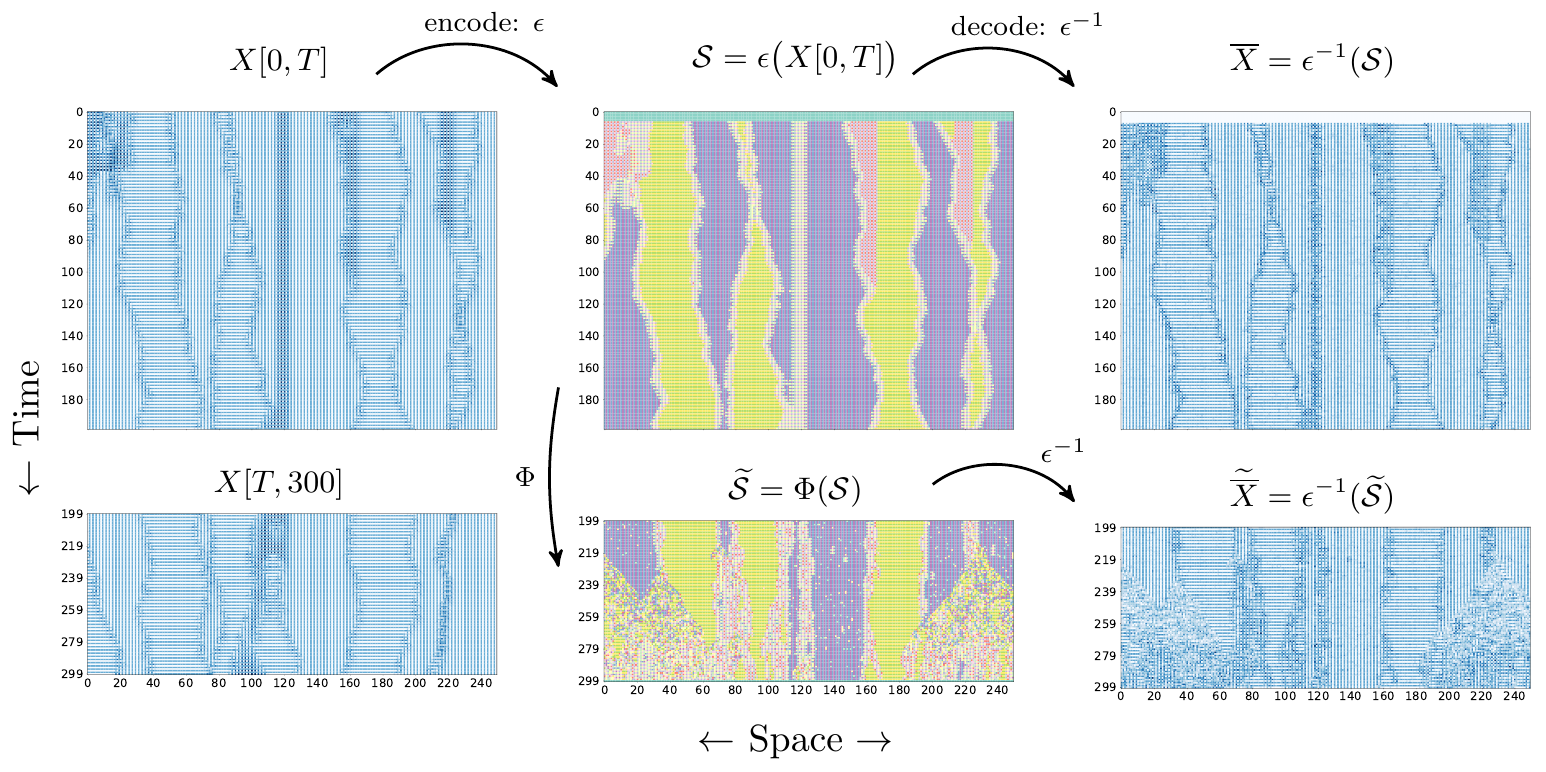}
\caption{Spacetime local causal state autoencoder: (Top-left) Observable spacetime field $X[0,T]$ of 	the circle map lattice used for training. A $250$-site portion of the $20,000$ spatial sites are shown (horizontal), which are evolved for an initial $200$ time steps (vertical) for training. (Top-middle) Encoded latent local causal state field $S=\epsilon\bigl(X[0,T]\bigr)$. Each unique color represents a unique local causal state. This compressed latent space provides a compressed representation of $X[0,T]$ for dimensionality-reduction. (Top-right) Reconstructed observable field $\overline{X} = \epsilon^{-1}(S)$, decoded from $S$. (Bottom-middle) Forecasted local causal state field, evolved forward in time after inference, starting at $T=199$, using $\widetilde{S} = \Phi(S)$ as described above. (Bottom-right) Forecasted observable field $\widetilde{\overline{X}}$, decoded from the evolved local causal state field $\widetilde{\overline{X}} = \epsilon^{-1}(\widetilde{S})$. (Bottom-left) Ground-truth observable field $X[T,300]$, evolved using the circle map lattice equation of motion.
}
\label{lcs-autoencoder}
\end{figure*}

Combining the $\epsilon$-function encoding with the $\epsilon^{-1}$-function decoding, the local causal states form a \emph{spacetime autoencoder}. As shown in the top portion of
Fig.~\ref{lcs-autoencoder}, the $\epsilon$-function encodes
an observable spacetime field $X$ to the compressed latent field $S$ and
$\epsilon^{-1}$ decodes to a reconstructed observable field $\overline{X} =
\epsilon^{-1}(S)$. Said another way, the identity $I \approx \epsilon \circ
\epsilon^{-1}$ is learned through the causal state bottleneck $S$.

There are several points to emphasize. First, unlike neural network
autoencoders, the local causal states are nonparametric models. And so, rather
than using the encoding and decoding together to train parameters as in neural
network autoencoders, the $\epsilon$-map and its inverse are learned directly
by approximating the local causal equivalence relation from data. Second, as
already noted, a crucial distinction is that the $\epsilon$-map encoding is
done locally so that the latent space and observable space share spacetime
coordinate geometry. The latent space of neural network autoencoders, in
contrast, does not share geometry with its inputs due to how their bottleneck
is created. For local causal states approximated from real-valued spacetime
data, the bottleneck comes from having a finite number of latent local causal
states and the latter are determined by the inherent structural dynamics.
However, this alone is a rather weak notion of autoencoder. Last, along these lines,
accounting for temporal evolution is critical for capturing pattern and
structure that is spontaneously generated by complex spatiotemporal systems.
Accounting for dynamics leads to a more powerful view of a causal state
autoencoder.

\subsection{Latent Space Dynamics}

In fact, a stochastic dynamic can be defined over the local causal states. This
combined with $\epsilon^{-1}$ decoding gives spacetime forecasting: infer the
local causal states and their dynamics up to the present time, evolve the
states forward in time, then decode to a forecasted observable field. This view
of the local causal states as \emph{predictive spacetime autoencoders} is much
more useful than the weak notion above. It synthesizes the two main aspects of
representation learning in physics---dimensionality-reduction and generative
modeling. For nonlinear dynamical systems the generative modeling of interest
is predictive forecasting, which generates sequential states and captures how
they are correlated and organized in time.

The local causal state field exhibits \emph{Markov shielding} \cite{Shal03a}.
The $\epsilon$-function uniquely determines a local causal state $\causalstate
= S\point$ from a full past lightcone $\plc\point$ of observables. If the local
causal states are known for each point in $\plc\point$, only the states in the
immediately preceding time step are required to determine $\causalstate$. That
is, $S\point$ is independent of the local causal states in $\plc\point$, given
the local causal states in its depth-$1$ past lightcone---the \emph{local
neighborhood} $\neighborhood \bigl(S(\site, t-1)\bigr)$ of $S(\site, t-1)$.

Markov shielding was derived in the setting of stochastic field theories, where
the observable field $X$ is \emph{not} from a deterministic dynamical system.
The stochasticity inherent in the system, along with Markov shielding, implies
the temporal dynamics over the local causal states is a \emph{stochastic
cellular automaton} (SCA), where $S\point$ is given by a stochastic function
$\phi$ of the local neighborhood $\neighborhood\bigl(S(\site, t-1) \bigr)$.

While our ultimate interests lie in deterministic systems (e.g., those governed
by partial differential equations), the local causal state dynamics is still an
SCA in this case. For systems with finite-range local interactions, which
\emph{define} a system's lightcones, a site value $X\point$ in $X$ is uniquely
determined by the past lightcone of $X\point$. In fact, for deterministic,
non-delay dynamics $X\point$ is uniquely determined by $\neighborhood
\bigl(X(\site, t-1) \bigr)$. In this way, Markov shielding in the local causal
states is inherited from the dynamics of the observable field. Since the local
causal states are compressed (local) representations, the local causal states
alone are not sufficient for deterministic evolution and, in general, the
information loss in the bottleneck implies the dynamic $\Phi$ over the local
causal states must be stochastic. Note that this argument is agnostic to the
functional form of the observable dynamics (i.e., the equations of motion). In
this way, the local causal states and their SCA dynamic $\Phi$ provide a
universal probabilistic model for spatially-extended dynamical systems.

Consider a finite set $\CausalStateSet$ of symbols (e.g., local causal state
labels). A radius-$\radius$ \emph{deterministic} cellular automata (CA) over
$\CausalStateSet$ is specified by a \emph{local update rule} $\phi:
\neighborhood \mapsto \causalstate$ that is a deterministic function of
radius-$\radius$ neighborhoods $\neighborhood$. In $1+1$-dimensions (one space,
one time), the neighborhoods are tuples of symbols from $\CausalStateSet$. For
example, in a radius-1 CA in $1+1$-dimensions: $\causalstate(\site, t+1) = \phi
\bigl(\neighborhood (\causalstate\point) \bigr) = \phi
\bigl(\causalstate(\site-1, t), \causalstate(\site, t) \causalstate(\site +1,
t) \bigr)$. The global CA dynamic $\Phi$, that evolves spatial configurations
over a time step, applies the local update $\phi$ synchronously and in parallel
across a configuration.  For the more general stochastic CAs, the local update
$\phi$ is still a function of local neighborhoods $\neighborhood$. However,
instead of outputting a single symbol $\causalstate$, it outputs a
\emph{probability mass function} (PMF) over symbols from $\CausalStateSet$.
Deterministic CAs then are the special case when the PMFs assign unity
mass to and only one symbol.

For a given observable spacetime field $X$, once the associated local causal
state field $S = \epsilon(X)$ has been inferred, the state dynamic $\Phi$ can
be estimated from $S$, again by simple counting. Empty histograms are
initialized for all possible neighborhoods $\neighborhood$ of local causal
states. For a particular neighborhood $\neighborhood_i$ found at $S\point$, the
histogram of $\phi(\neighborhood_i)$ is incremented at $\causalstate$ for
$\causalstate = S(\site, t+1)$.

Since real-valued systems require approximation schemes and models are always
inferred from finite data, using $\Phi$ estimated in this way to evolve the
local causal states forward in time may yield a neighborhood $\neighborhood$
not seen in the original inferred field $S$. The local dynamic $\phi$ outputs
an empty histogram in this case. In practice, to circumvent this issue we keep
and update a separate PMF over local causal states that is the spatial
distribution over the local causal states.

At each  $t$ we estimate a histogram over the local causal states according to
the spatial configuration $S(t)$ of the local causal state field. If an empty
histogram is encountered during the evolution $S(t+1) = \Phi \bigl( S(t)
\bigr)$, a local causal state for that point is chosen from the spatial PMF
instead. This heuristic leaves room for variation, and we use a
running-estimate spatial distribution since the distribution over local causal
states generally is not temporally stationary.

\section{Forecasting Spacetime Fields}

To demonstrate that local causal states are spacetime autoencoders, along with
their predictive forecasting, the following presents preliminary results for a
nonlinear map lattice system based on the \emph{circle map}. While the choice
is somewhat arbitrary, we selected the circle map lattice due to the
complexity of the self-organized patterns and structures it generates. Map
lattices are also markedly simpler to simulate than PDEs, such as the
Kuramoto-Sivashinsky equation popular in data-driven forecasting
explorations~\cite{Path18a,Otto19a}, that require elaborate numerical
integration schemes.

A one-dimensional \emph{map lattice} \cite{Crut87b} is a spatially-extended
dynamical system that evolves configurations on a discrete spatial lattice
$\mathbb{Z}$ in discrete time steps according to the local dynamics:
\begin{align}
\begin{split}
x(\site, t+1) & = (1 - \alpha)f\bigl(x\point \bigr) + \\
  &\frac{\alpha}{2} \bigl[f\bigl(x(\site+1, t) \bigr)
  + f\bigl(x(\site-1, t) \bigr) \bigr]
  ~,
  \end{split}
  \label{eqn:map-lattice}
\end{align}
where $\site$ is the spatial index, $t$ is the time index, $\alpha$ is the
coupling strength, and $f$ is an iterated map of the unit interval: $x(t+1) =
f\bigl(x(t)\bigr), ~ x \in [0,1]$. We use the circle map:
\begin{align*}
f(x) = x + \omega - \frac{K}{2\pi} \sin(2 \pi x) \; \mathrm{mod} \; 1
  ~,
\end{align*}
where $\omega$ is a phase shift and $K$ is the strength of the nonlinearity.
Following \cite{Grav18a}, Fig.~\ref{lcs-autoencoder} uses $\omega = 0.5$,
$K = 1.0$, and $\alpha = 1.0$.

The map lattice's complex behaviors seen there in the observable spacetime
field $X$ in the left panels arise from two competing background \emph{domains}.
One is spatial period-$2$ (vertical stripes) and the other is temporal
period-$2$ (horizontal stripes).  Interfaces between these domains (also know as \emph{dislocations} in the statistical mechanics literature) diffuse
through space over time and sometimes pairwise annihilate upon collision.
Due to pairwise annihilation, the macroscopic behavior of domains and the interactions of their interfaces is not stationary in time. 

When evolved from random initial conditions, as done here, there are initially
many domain interfaces  that quickly annihilate. These pairwise interactions
decrease over time until the system's spacetime behavior is mostly comprised of
domain regions. We are interested here in intermediate times that have a
balance between dislocation dynamics and (meta-)stable domain regions. To this
end, we let the map lattice evolve for $300$ time steps before starting local
causal state inference. Note that time $t=0$ in Fig.~\ref{lcs-autoencoder}
starts after this initial transient time. We use a lattice size of $N=20,000$
(only a portion of which is shown in Fig.~\ref{lcs-autoencoder}) and periodic
boundary conditions for our map lattice. 

After the initial $300$-step transient time, the map lattice is evolved for
another $T=200$ time steps to produce the training observable field $X[0,T]$
that is used for local causal state inference. For inference we use past
lightcone depth $h^- = 6$, future lightcone depth $h^+ = 1$, and propagation
speed $c = 1$ (determined by the radius $\radius = 1$ local interactions of the
map lattice). In the lightcone clustering step, i.e. $\gamma$-equivalence, we
use $K = 10$ for past lightcone clustering and $K = 40$ for future lightcone
clustering. For $\psi$-equivalence, we use hierarchical agglomerative
clustering, using a $\chi^2$ test with $\alpha = 0.05$ for distribution comparison. 

After inference, the training observable field $X[0,T]$ is encoded using the approximated $\epsilon$-map, $\epsilon(\plc) \approx \psi\bigl( \gamma^-(\plc) \bigr)$, to produce the associated local causal state field $S = \epsilon \bigl(X[0,T]\bigr)$, shown in the top-middle panel of Fig.~\ref{lcs-autoencoder}. Each unique local causal state is assigned an arbitrary integer label during inference, and these integer labels are then assigned arbitrary colors for the latent field visualizations in Fig.~\ref{lcs-autoencoder}. Each unique color seen in $S$ identifies a unique local causal state at that point.

As the two domains in $X$ are period-$2$, each is described by two local causal
states in $S = \epsilon(X)$. The result is that only four local causal states
capture the majority of the observed spacetime behavior, with additional local
causal states capturing particular interactions at the domain interfaces. From
the dimensionality-reduction perspective, these are the structures that one
would like to capture and in terms of which we would then re-express the
system's evolution. Note that at the ``microscopic'' level, in terms of the
full spatial lattice, the dynamics of the system are deterministic, following
Eq.~(\ref{eqn:map-lattice}), while the evolution of the reduced ``macroscopic''
description is probabilistic, following the stochastic diffusion and
annihilation of domain interfaces.

Recall that the $\epsilon^{-1}$-function is also learned during inference, and it is then used to create the reconstruction $\overline{X} = \epsilon^{-1}(S)$ of $X[0,T]$, shown in the top-right panel of Fig.~\ref{lcs-autoencoder}. The reconstructed observable field $\overline{X}$ qualitatively reproduces the macroscopic behavior observed in $X[0,T]$, even though the microscopic details are noisy due to the stochasticity of the decoding. Given that the latent space $S$ is comprised of just $10$ local causal states, the reconstructed observable field $\overline{X} = \epsilon^{-1}(S)$ is surprisingly faithful to the original observable field $X[0,T]$.  

Once the local causal state field $S$ has been inferred, the stochastic latent
space dynamic $\Phi$ can be estimated, as described above. Using the estimated
dynamic, the latent space is evolved forward in time to produce a forecasted
local causal state field $\widetilde{S} = \Phi(S)$. In the bottom-middle of
Fig.~\ref{lcs-autoencoder} the local causal state field is predicted forward in
time another $100$ time steps. Note that in the original inferred latent space
field $S = \epsilon\bigl(X[0,T]\bigr)$ no local causal states are placed in
regions that do not have full past or future lightcones of the chosen depths.
These points are known collectively as the \emph{margin} of the latent space
field $S$. Due to the periodic boundaries in space, there are only time
margins. Margin points are assigned a unique label in $S$, rather than just
being omitted, so that $S$ is the same shape as the observable field it is
trained on. Since we use future lightcone depth $h^+ = 1$ here, there are no local causal states assigned at time step $199$. Thus, as can be seen in the bottom panels of Fig.~\ref{lcs-autoencoder}, the forecast starts at $T = 199$. 

There are several points to note for the forecasted local causal state field
$\widetilde{S} = \Phi(S)$ in the bottom-middle panel. We again emphasize that
the following analysis is enabled and enhanced by the visual interpretability
afforded by the shared coordinate geometry of the observable and latent spaces.
From visual inspection, we can see that the most stable and well-predicted
regions are the periodic domains. By relying on the previous-time spatial
distribution when null PMFs are encountered applying $\Phi$, the domain
regions are stable to incorrect predictions, to some degree. We can see this,
for instance, in the predicted domain region around site $r = 140$. There is
noise present in the prediction of this domain, as there are local causal
states present other than the two states that identify this domain in the same
region in the inferred field $S = \epsilon \bigl(X[0,T]\bigr)$ just before the
forecast starts. The domain persists in the prediction, despite this noise. 

However, resilience to incorrect predictions does not hold generally. As can be
seen, for instance on the right-hand side of $\widetilde{S}$ after $t \approx
220$, prediction errors can sometimes cause cascading failures in the forecast,
which typically travel around the propagation speed of $c = 1$. Once a
cascading failure starts, it is generally not recoverable due to instability in the inferred dynamic and the forecast breaks down. 

Again using the $\epsilon^{-1}$-function, we decode $\widetilde{S}$ to produce
a forecasted observable field $\widetilde{\overline{X}} =
\epsilon^{-1}(\widetilde{S})$. As with the reconstructed field $\overline{X}$,
the small-scale details are noisy, but the large-scale behavior predicted by
the local causal state field $\widetilde{S}$ is reproduced in the observable
forecast. Comparing to the ground-truth observable field $X[T, 300]$,
provided by evolving the map lattice forward another $100$ time steps after
creating the training field, we see that the domain regions are forecasted
well. Several interfaces remain stable for the full duration without breaking
down into a cascading failure, such as those near $r = 140$. 

Ideally, the forecast should not produce the kind of cascading failures seen in
$\widetilde{S}$ and $\widetilde{\overline{X}}$. These are likely due to empty
PMFs encountered when applying the estimated latent space dynamic $\Phi$.
Clearly, there is room for improvement in the algorithms, heuristics, and
training protocols. These preliminary results nonetheless demonstrate the
two-fold benefit of the local causal states for physics-based representation
learning: dimensionality-reduction and generative modeling---predictive
forecasting, in this case. However the method is improved, we
suspect there will ultimately be a trade-off between these two aspects of
local causal states. Allowing for more local causal states to be learned during
inference (the hyperparameter choice of $K$ in K-Means cluster over past
lightcones sets an upper bound to the possible number of states that can be
learned) will generally increase forecasting performance, but at the same time
detract from with visual interpretability of the latent space for
dimensionality-reduction. 

Python source code and a Jupiter notebook that produces the results shown in Fig.~\ref{lcs-autoencoder} is available at \url{https://github.com/adamrupe/spacetime_autoencoders}.

\paragraph*{Acknowledgments}
AR and JPC acknowledge Intel's support for the IPCC at UC Davis and thank the
Telluride Science Research Center for its hospitality during visits. This
research is based upon work supported by, or in part by, the U. S. Army
Research Laboratory and the U. S. Army Research Office under contract
W911NF-18-1-0028.
Part of this research was performed while AR was visiting the Institute for Pure and Applied Mathematics (IPAM), which is supported by the National Science Foundation (Grant No. DMS-1440415).

\bibliographystyle{aaai}
\bibliography{spacetime,chaos}

\end{document}